\documentclass{article} % For LaTeX2e
\usepackage{iclr2017_conference,times}
\usepackage{hyperref}
\usepackage{url}
\usepackage[pdftex]{graphicx}
\usepackage{amsmath,amssymb} % define this before the line numbering.
\usepackage{subfig}
\usepackage{multirow}

\title{Energy-Based Spherical Sparse Coding}

\author{Bailey Kong and Charless C. Fowlkes \\
Department of Computer Science \\
University of California, Irvine \\
Irvine, CA 92697 USA \\
\texttt{\{bhkong,fowlkes\}@ics.uci.edu}
}

\renewcommand\vec[1]{\mathbf{#1}}

\newcommand\symvec[1]{\boldsymbol{#1}}

\newcommand\mat[1]{\mathbf{#1}}

\newcommand\transpose[1]{#1^\intercal}

\renewcommand\L{\mathcal{L}}

\newcommand\R{\mathbb{R}}

\newcommand\eg{e.g., }
\newcommand\ie{i.e., }
\newcommand\wrt{w.r.t. }

\iclrfinalcopy % Uncomment for camera-ready version

\begin{document}

\maketitle

\begin{abstract}
In this paper, we explore an efficient variant of convolutional sparse coding
with unit norm code vectors where reconstruction quality is evaluated using an
inner product (cosine distance). To use these codes for discriminative
classification, we describe a model we term Energy-Based Spherical Sparse
Coding (EB-SSC) in which the hypothesized class label introduces a learned
linear bias into the coding step.  We evaluate and visualize performance of
stacking this encoder to make a deep layered model for image classification.
\footnote{This work was supported by NSF grants DBI-1262574, IIS-1253538, and a
hardware donation from NVIDIA.}
\end{abstract}

\section{Introduction}

Sparse coding has been widely studied as a representation for images, audio and
other vectorial data.  This highly successful method that has found
its way into many applications, from signal compression and
denoising~\citep{donoho2006compressed,elad2006image} to image
classification~\citep{wright2009robust}, to modeling
neuronal receptive fields in visual cortex~\citep{olshausen97}.
Since its introduction, subsequent works have brought sparse coding into the
supervised learning setting by introducing classification loss terms to the
original formulation to encourage features that are not only able to
reconstruct the original signal but are also discriminative 
~\citep{LC-KSVD,yang_supervised_2010,zeiler_deconvolutional_2010,ji_hierarchical_2011,zhou2012learning,zhang_discriminative_2013}.

While supervised sparse coding methods have been shown to find more discriminative
features leading to improved classification performance over their unsupervised
counterparts, they have received much less attention in recent years and have
been eclipsed by simpler feed-forward architectures.

This is in part because sparse coding is computationally expensive.  Convex
formulations of sparse coding typically consist of a minimization problem
over an objective that includes a least-squares (LSQ) reconstruction error term
plus a sparsity inducing regularizer.

Because there is no closed-form solution to this formulation, various iterative
optimization techniques are generally used to find a
solution~\citep{zeiler_deconvolutional_2010,bristow_fast_2013,yang2013fast,heide_fast_2015}.
In applications where an approximate solution suffices, there is work that
learns non-linear predictors to estimate sparse codes rather than solve the
objective more directly~\citep{gregor2010learning}.  The computational overhead
for iterative schemes becomes quite significant when training discriminative
models due to the demand of processing many training examples necessary for
good performance, and so sparse coding has fallen out of favor by not being
able to keep up with simpler non-iterative coding methods.

%(2) Euclidean distance is not a good metric for many features in computer vision
%applications~\citep{yan2007exploring,wu2009beyond,choi2014toward}.
%This problem is frequently attributed to that of the \emph{curse of
%dimensionality}, where the proportional difference between the distance of the
%furthest-points and the closest-points vanish as the dimensionality
%increases~\citep{beyer1999nearest}.
%Contrary to popular thinking, \citet{houle2010can} suggest the actual problem to
%be that of irrelevant features.
%Since Euclidean distance weighs all dimensions equally, irrelevant features
%are emphasized just as much as relevant features are.

\begin{figure}[h]
\centering
\subfloat[CReLU]{%
  \begin{minipage}[c][1.05\width]{0.33\textwidth}
    \centering
    \includegraphics[width=0.93in]{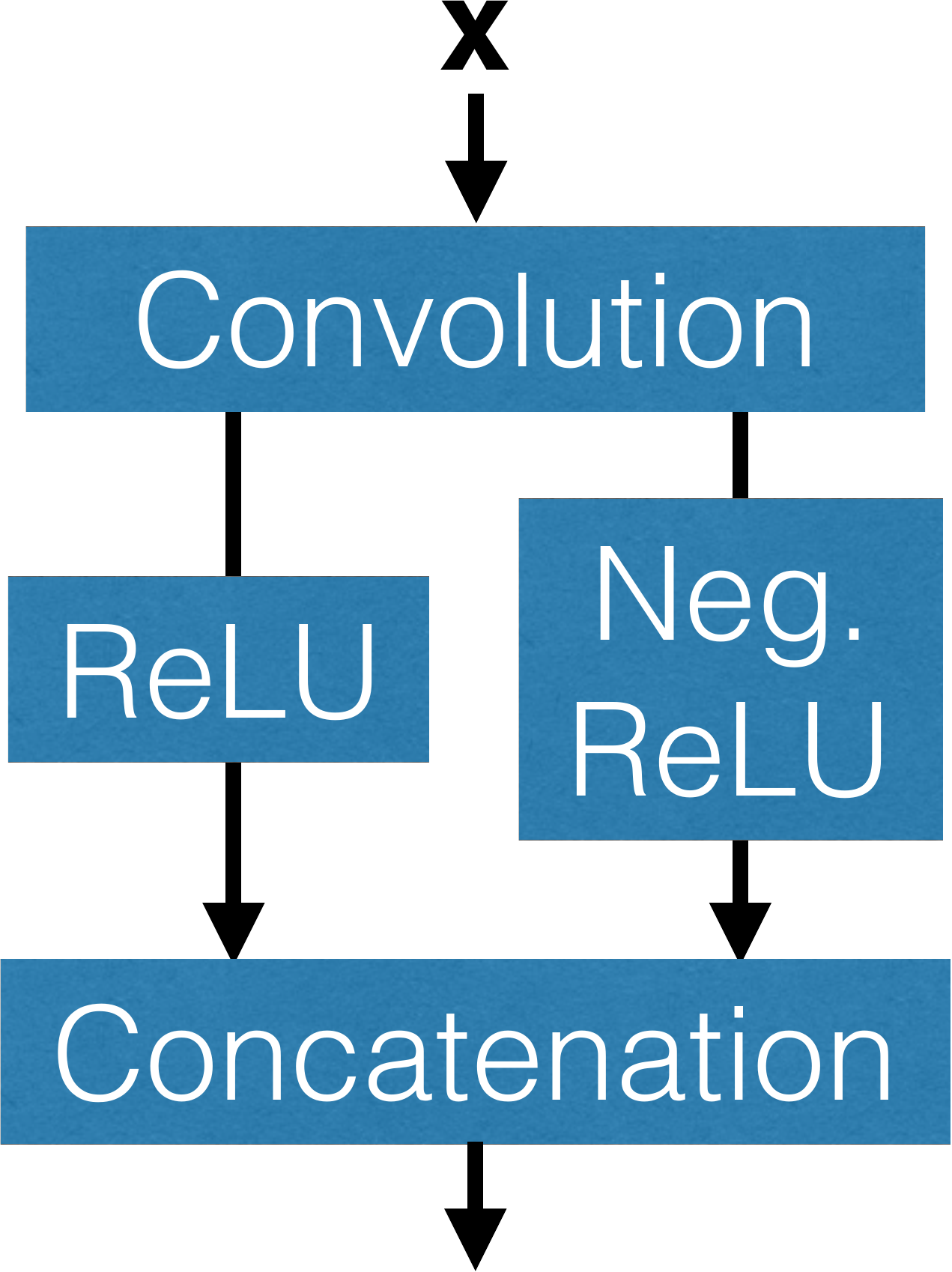}
  \end{minipage}}
\subfloat[SSC]{%
  \begin{minipage}[c][1.05\width]{0.33\textwidth}
    \centering
    \includegraphics[width=0.93in]{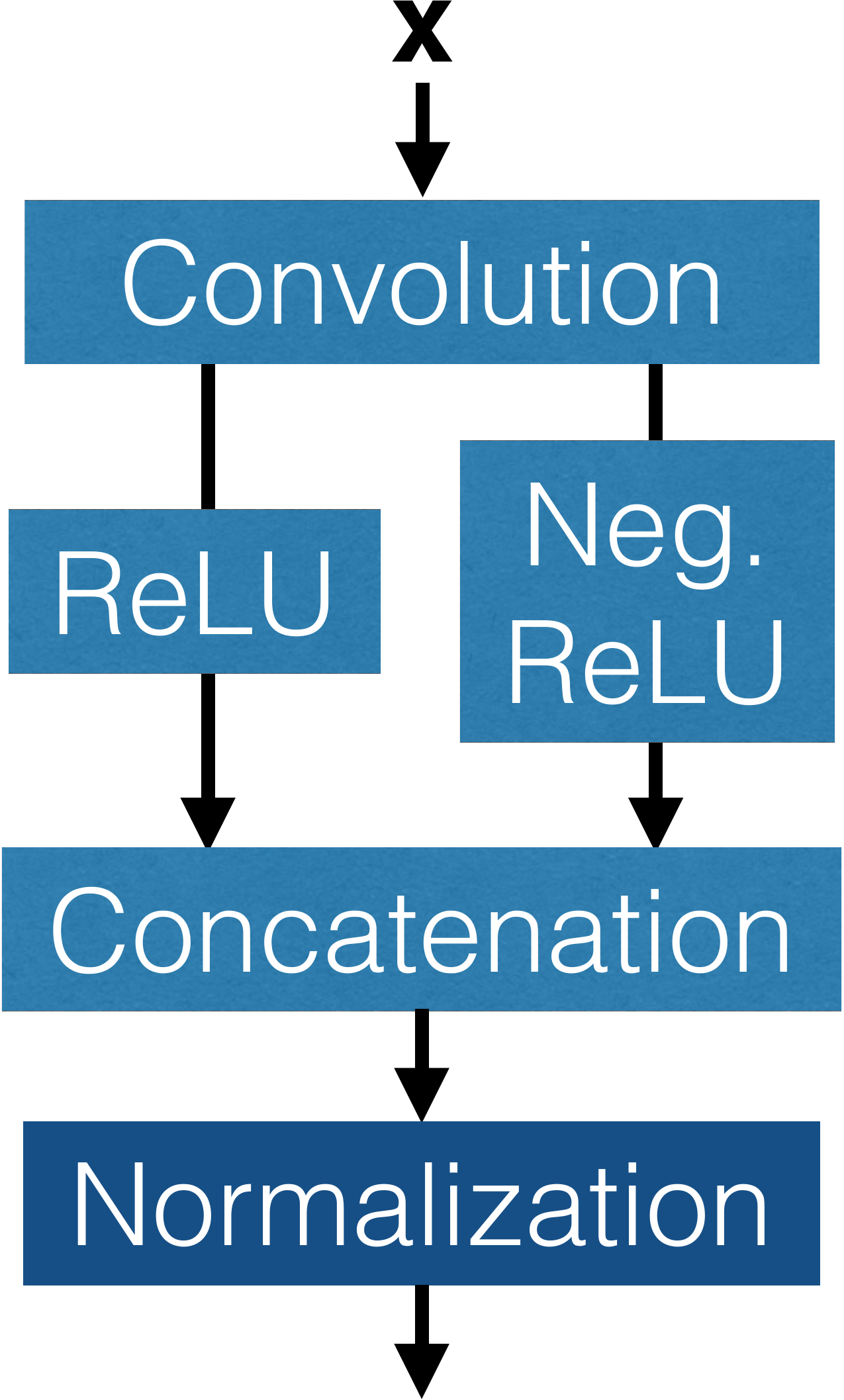}
  \end{minipage}}
\subfloat[EB-SSC]{%
  \begin{minipage}[c][1.05\width]{0.33\textwidth}
    \centering
    \includegraphics[width=1.3in]{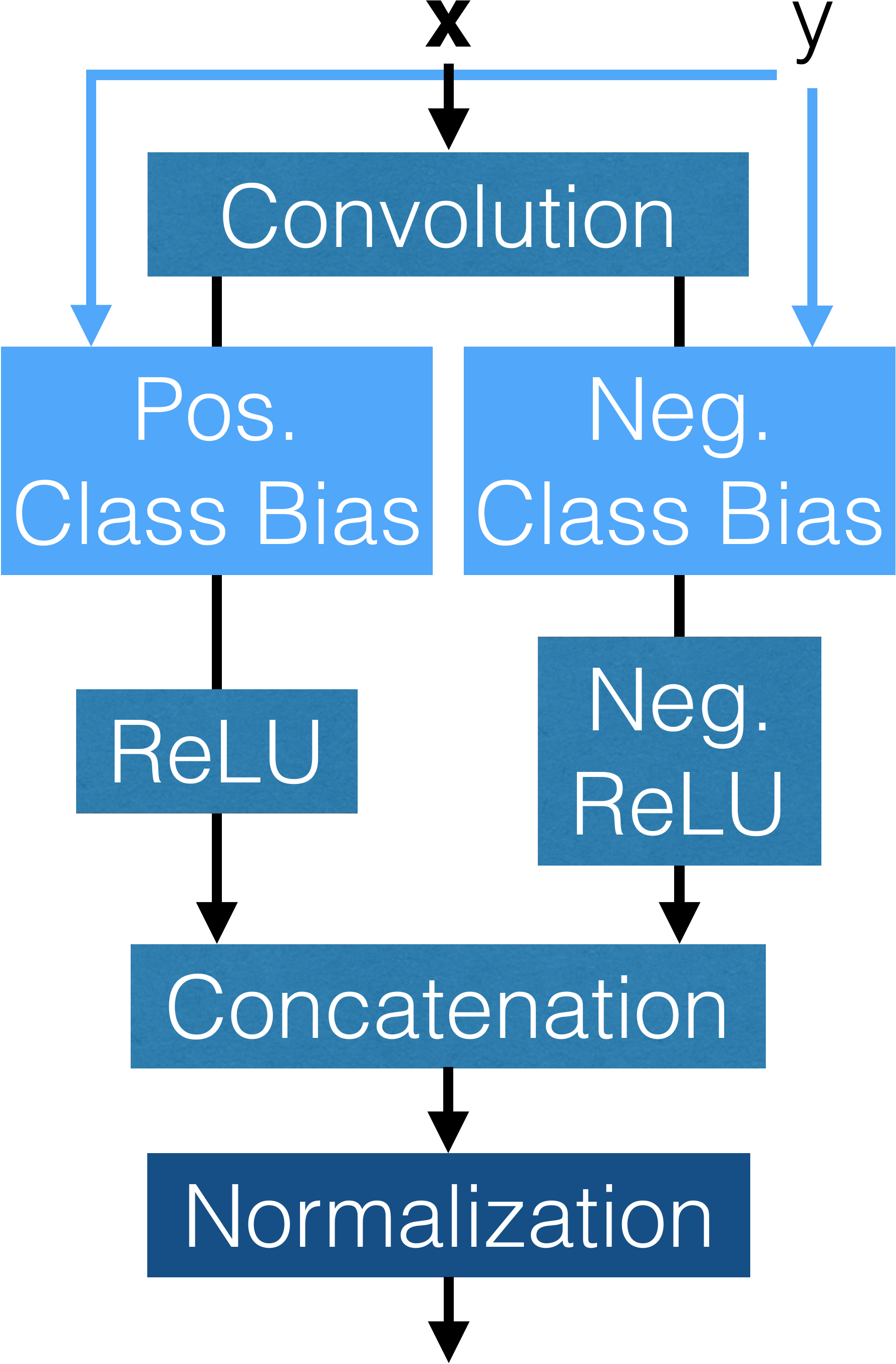}
  \end{minipage}}
\caption{Building blocks for coding networks explored in this paper. Our 
coding model uses non-linearities that are closely related to the standard
ReLU activation function.  (a) Keeping both positive and negative activations
provides a baseline feed-forward model termed concatenated ReLU (CReLU).
(b) Our spherical sparse coding layer has a similar structure but with an
extra bias and normalization step.  Our proposed energy-based model uses (c)
energy-based spherical sparse coding (EB-SSC) blocks that produces sparse
activations which are not only positive and negative, but are class-specific.
These blocks can be stacked to build deeper architectures.}
\label{fig:ebm_nn}
\end{figure}

In this paper we introduce an alternate formulation of sparse coding using unit
length codes and a reconstruction loss based on the cosine similarity. Optimal
sparse codes in this model can be computed in a non-iterative fashion and the
coding objective lends itself naturally to embedding in a discriminative,
energy-based classifier which we term \emph{energy-based spherical sparse
coding (EB-SSC)}.  This bi-directional coding method incorporates both top-down
and bottom-up information where the features representation depends on both a
hypothesized class label and the input signal.  Like \citet{cao2015look}, our
motivation for bi-directional coding comes from the ``Biased Competition
Theory'', which suggests that visual processing can be biased by other mental
processes (\eg top-down influence) to prioritize certain features that are most
relevant to current task.  Fig.~\ref{fig:ebm_nn} illustrates the flow of
computation used by our SSC and EB-SSC building blocks compared to a standard
feed-forward layer.

Our energy based approach for combining top-down and bottom-up information is
closely tied to the ideas of
\citet{larochelle_classification_2008,ji_hierarchical_2011,zhang_discriminative_2013,li_bi-directional_2014}---although
the model details are substantially different (\eg \citet{ji_hierarchical_2011}
and \citet{zhang_discriminative_2013} use sigmoid non-linearities while
\citet{li_bi-directional_2014} use separate representations for top-down and
bottom-up information).
The energy function of \citet{larochelle_classification_2008} is also similar
but includes an extra classification term and is trained as a restricted
Boltzmann machine.

\subsection{Notation}
Matrices are denoted as uppercase bold (\eg $\mat{A}$), vectors are lowercase
bold (\eg $\vec{a}$), and scalars are lowercase (\eg $a$).
We denote the transpose operator with $^\intercal$, the element-wise multiplication
operator with $\odot$, the convolution operator with $*$, and the
cross-correlation operator with $\star$.
For vectors where we dropped the subscript $k$ (\eg $\vec{d}$ and $\vec{z}$),
we refer to a super vector with $K$ components stacked together (\eg
$\vec{z} = \transpose{[\transpose{\vec{z}_1},\ldots,\transpose{\vec{z}_K}] }$).

\section{Energy-Based Spherical Sparse Coding}
Energy-based models capture dependencies between variables using an energy
function that measure the compatibility of the configuration of
variables~\citep{lecun2006tutorial}.
To measure the compatibility between the top-down and bottom-up information, we
define the energy function of EB-SSC to be the sum of bottom-up coding term
and a top-down classification term:
\begin{equation}
E(\vec{x},y,\vec{z}) = E_{code}(\vec{x},\vec{z})+E_{class}(y,\vec{z}).
\label{eq:energy_fn}
\end{equation}
The bottom-up information (input signal $\vec{x}$) and the top-down information (class
label $y$) are tied together by a latent feature map $\vec{z}$.

\subsection{Bottom-Up Reconstruction}
To measure the compatibility between the input signal $\vec{x}$ and the latent feature
maps $\vec{z}$, we introduce a novel variant of sparse coding that is amenable
to efficient feed-forward optimization.
While the idea behind this variant can be applied to either patch-based or
convolutional sparse coding, we specifically use the convolutional variant that
shares the burden of coding an image among nearby overlapping dictionary
elements.
Using such a shift-invariant approach avoids the need to learn dictionary
elements which are simply translated copies of each other, freeing up resources
to discover more diverse and specific filters (see
\citet{kavukcuoglu_learning_2010}).

Convolutional sparse coding (CSC) attempts to find a set of dictionary elements
$\{\vec{d}_1,\ldots,\vec{d}_K\}$ and corresponding sparse codes
$\{\vec{z}_1,\ldots,\vec{z}_K\}$ so that the resulting reconstruction,
$\vec{r}=\sum_{k=1}^K \vec{d}_k*\vec{z}_k$ accurately represents the input
signal $\vec{x}$.
This is traditionally framed as a least-squares minimization with a sparsity
inducing prior on $\vec{z}$:
\begin{equation}
  \underset{\vec{z}}{\arg\min}\,\,
    \|\vec{x}-\sum_{k=1}^K\vec{d}_k*\vec{z}_k\|_2^2 +
    \beta\|\vec{z}\|_1.
  \label{eqn:lsqerror}
\end{equation}
Unlike standard feed-forward CNN models that convolve the input signal $\vec{x}$
with the filters, this energy function corresponds to a generative model where
the latent feature maps $\{\vec{z}_1,\ldots,\vec{z}_K\}$ are convolved with the
filters and compared to the input
signal~\citep{bristow_fast_2013,heide_fast_2015,zeiler_deconvolutional_2010}.

To motivate our novel variant of CSC, consider expanding the squared
reconstruction error $\|\vec{x} - \vec{r}\|^2_2 = \|\vec{x}\|_2^2 -
2\transpose{\vec{x}}\vec{r} + \|\vec{r}\|_2^2$.
If we constrain the reconstruction $\vec{r}$ to have unit norm, the
reconstruction error depends entirely on the inner product between $\vec{x}$ and
$\vec{r}$ and is equivalent to the cosine similarity (up to additive and
multiplicative constants).
This suggests the closely related \emph{unit-length} reconstruction problem:
\begin{align}
  \underset{\vec{z}}{\arg\max}\,\, &
    \transpose{\vec{x}}\big(\sum_{k=1}^K\vec{d}_k*\vec{z}_k\big) -
    \beta\|\vec{z}\|_1 \label{eqn:iperror} \\
  \text{s.t. }\,\, &\big\|\sum_{k=1}^K\vec{d}_k*\vec{z}_k\big\|_2 \leq 1 \nonumber
\end{align}
In \nameref{app:deriv_ipcsc} we show that, given an optimal unit length
reconstruction $\bar{\vec{r}}^*$ with corresponding codes $\bar{\vec{z}}^*$,
the solution to the least squares reconstruction problem
(Eq.~\ref{eqn:lsqerror}) can be computed by a simple scaling $\vec{r}^* =
(\transpose{\vec{x}}
\bar{\vec{r}^*}-\frac{\beta}{2}\|\bar{\vec{z}}^*\|_1)\bar{\vec{r}}^*$.

The unit-length reconstruction problem is no easier than the original
least-squares optimization due to the constraint on the reconstruction which
couples the codes for different filters.  Instead consider a simplified
constraint on $\vec{z}$ which we refer to as \emph{spherical sparse coding
(SSC)}:
\begin{equation}
  \underset{\|\vec{z}\|_2 \leq 1}{\arg\max}\,\,
    E_{code}(\vec{x},\vec{z}) =
  \underset{\|\vec{z}\|_2 \leq 1}{\arg\max}\,\,
    \transpose{\vec{x}}\big( \sum_{k=1}^K \vec{d}_k * \vec{z}_k \big) -
    \beta\|\vec{z}\|_1.
    \label{eq:energy_btup}
\end{equation}
In \ref{sec:coding} below, we show that the solution to this problem can be
found very efficiently without requiring iterative optimization.

This problem is a relaxation of convolutional sparse coding since it ignores
non-orthogonal interactions between the dictionary elements\footnote{We note
that our formulation is also closely related to the dynamical model suggested
by \citet{rozell_sparse_2008}, but without the dictionary-dependent lateral
inhibition between feature maps.  Lateral inhibition can solve the unit-length
reconstruction formulation of standard sparse coding but requires iterative
optimization.}.
Alternately, assuming unit norm dictionary elements, the code norm constraint
can be used to upper-bound the reconstruction length.  We have by the triangle
and Young's inequality that:
\begin{eqnarray}
  \big\|\sum_k \vec{d}_k * \vec{z}_k\big\|_2
  \leq \sum_k \| \vec{d}_k * \vec{z}_k\|_2
  \leq \sum_k \| \vec{d}_k\|_1 \|\vec{z}_k\|_1
  \leq D \sum_k \|\vec{z}_k\|_2
\end{eqnarray}
where the factor $D$ is the dimension of $\vec{z}_k$ and arises from switching
from the $1$-norm to the $2$-norm.  Since $D \sum_k \|\vec{z}_k\|_2 \leq 1$
is a tighter constraint we have
\begin{equation}
  \underset{\|\sum_k \vec{d}_k*\vec{z}_k\|_2 \leq 1}{\max}\,\, E_{code}(\vec{x},\vec{z}) 
  \geq
  \underset{\sum_k \|\vec{z}_k\|_2 \leq \frac{1}{D}}{\max}\,\, E_{code}(\vec{x},\vec{z})
%  = \underset{\sum_k \|\vec{z}_k\|_2 \leq 1}{\max}\,\, D^{-1} \cdot E_{code}(\vec{x},\vec{z})  
\end{equation}
%Since our objective scales linearly in the norm of $\vec{z}$,  we 
%  \underset{\|\vec{z}_k\|_2 \leq 1}{\max}\,\, D^{-1}\cdot E_{code}(\vec{x},\vec{z}) =

However, this relaxation is very loose, primarily due to the triangle inequality.
Except in special cases (\eg if the dictionary elements have disjoint spectra)
the SSC codes will be quite different from the standard least-squares
reconstruction.

\subsection{Top-Down Classification}
To measure the compatibility between the class label $y$ and the latent feature
maps $\vec{z}$, we use a set of one-vs-all linear classifiers.  To provide more
flexibility, we generalize this by splitting the code vector into positive and
negative components:
\begin{equation*}
  \vec{z}_k = \vec{z}_k^+ + \vec{z}_k^-
  \quad
  \vec{z}_k^+ \geq 0 \quad \vec{z}_k^- \leq 0
\end{equation*}
and allow the linear classifier to operate on each component separately.  We
express the classifier score for a hypothesized class label $y$ by:
\begin{equation}
  E_{class}(y,\vec{z}) =
    \sum_{k=1}^K \vec{w}_y^{+\intercal}\vec{z}_k^+ +
    \sum_{k=1}^K \vec{w}_y^{-\intercal}\vec{z}_k^-.
  \label{eq:energy_tpdw}
\end{equation}
The classifier thus is parameterized by a pair of weight vectors
($\vec{w}_{yk}^+$ and $\vec{w}_{yk}^-$) for each class label $y$ and $k$-th
channel of the latent feature map.

This splitting, sometimes referred to as full-wave rectification, is useful
since a dictionary element and its negative do not necessarily have opposite
visual semantics.  This splitting also allows the classifier the flexibility to
assign distinct meanings or alternately be completely invariant to contrast
reversal depending on the problem domain.  For example,
\citet{shang2016understanding} found CNN models with ReLU non-linearities which
discard the negative activations tend to learn pairs of filters which are
related by negation.  Keeping both positive and negative responses allowed them
to halve the number of dictionary elements.  

We note that it is also straightforward to introduce spatial average pooling
prior to classification by introducing a fixed linear operator $\mat{P}$ used
to pool the codes (e.g., $\vec{w}_y^{+\intercal}\mat{P}\vec{z}_k^+$).  This is
motivated by a variety of hand-engineered feature extractors and sparse coding
models, such as \citet{ren_histograms_2013}, which use spatially pooled
histograms of sparse codes for classification.  This fixed pooling can be
viewed as a form of regularization on the linear classifier which enforces
shared weights over spatial blocks of the latent feature map.  Splitting is
also quite important to prevent information loss when performing additive
pooling since positive and negative components of $\vec{z}_k$ can cancel each
other out.

\subsection{Coding}
\label{sec:coding}
Bottom-up reconstruction and top-down classification each provide half of the
story, coupled by the latent feature maps.
For a given input $\vec{x}$ and hypothesized class $y$, we would like to find
the optimal activations $\vec{z}$ that maximize the joint energy function
$E(\vec{x},y,\vec{z})$.
This requires solving the following optimization:
\begin{equation}
  \underset{\|\vec{z}\|_2\le1}{\arg\max}\,\,
    \transpose{\vec{x}}\big( \sum_{k=1}^K \vec{d}_k*\vec{z}_k \big)
    - \beta\|\vec{z}\|_1
    + \sum_{k=1}^K \vec{w}_{yk}^{+\intercal}\vec{z}_k^+
    + \sum_{k=1}^K \vec{w}_{yk}^{-\intercal}\vec{z}_k^-,
    \label{eq:code_energy}
\end{equation}
where $\vec{x}\in\R^D$ is an image and $y\in\mathcal{Y}$ is a class hypothesis.
$\vec{z}_k\in\R^F$ is the $k$-th component latent variable being inferred;
$\vec{z}_k^+$ and $\vec{z}_k^-$ are the positive and negative coefficients of
$\vec{z}_k$, such that $\vec{z}_k = \vec{z}_k^+ + \vec{z}_k^-$.
The parameters $\vec{d}_k\in\R^M$, $\vec{w}_{yk}^+\in\R^F$, and
$\vec{w}_{yk}^-\in\R^F$ are the dictionary filter, positive coefficient
classifier, and negative coefficient classifier for the $k$-th component
respectively.
A key aspect of our formulation is that the optimal codes can be found very
efficiently in closed-form---in a feed-forward manner (see \nameref{app:coding_ipcsc}
for a detailed argument).

\subsubsection{Asymmetric Shrinkage}
\label{sec:asym}
To describe the coding processes, let us first define a generalized version of
the shrinkage function commonly used in sparse coding.
Our asymmetric shrinkage is parameterized by upper and lower thresholds
$-\beta^- \leq \beta^+$
\begin{align}
  \operatorname{shrink}_{(\beta^+,\beta^-)}(v) &= \left\{
    \begin{array}{cl}
      v-\beta^+ & \qquad \text{if } v-\beta^+>0 \\
      0 & \qquad \text{otherwise} \\
      v+\beta^- & \qquad \text{if } v+\beta^-<0
    \end{array}
    \right.
    \label{eq:asym_shrink}
\end{align}

\begin{figure}[h]
\centering
\begin{tabular}{cccc}
  \subfloat[$-\beta^- \le 0 \le \beta^+$]{\includegraphics[height=3cm]{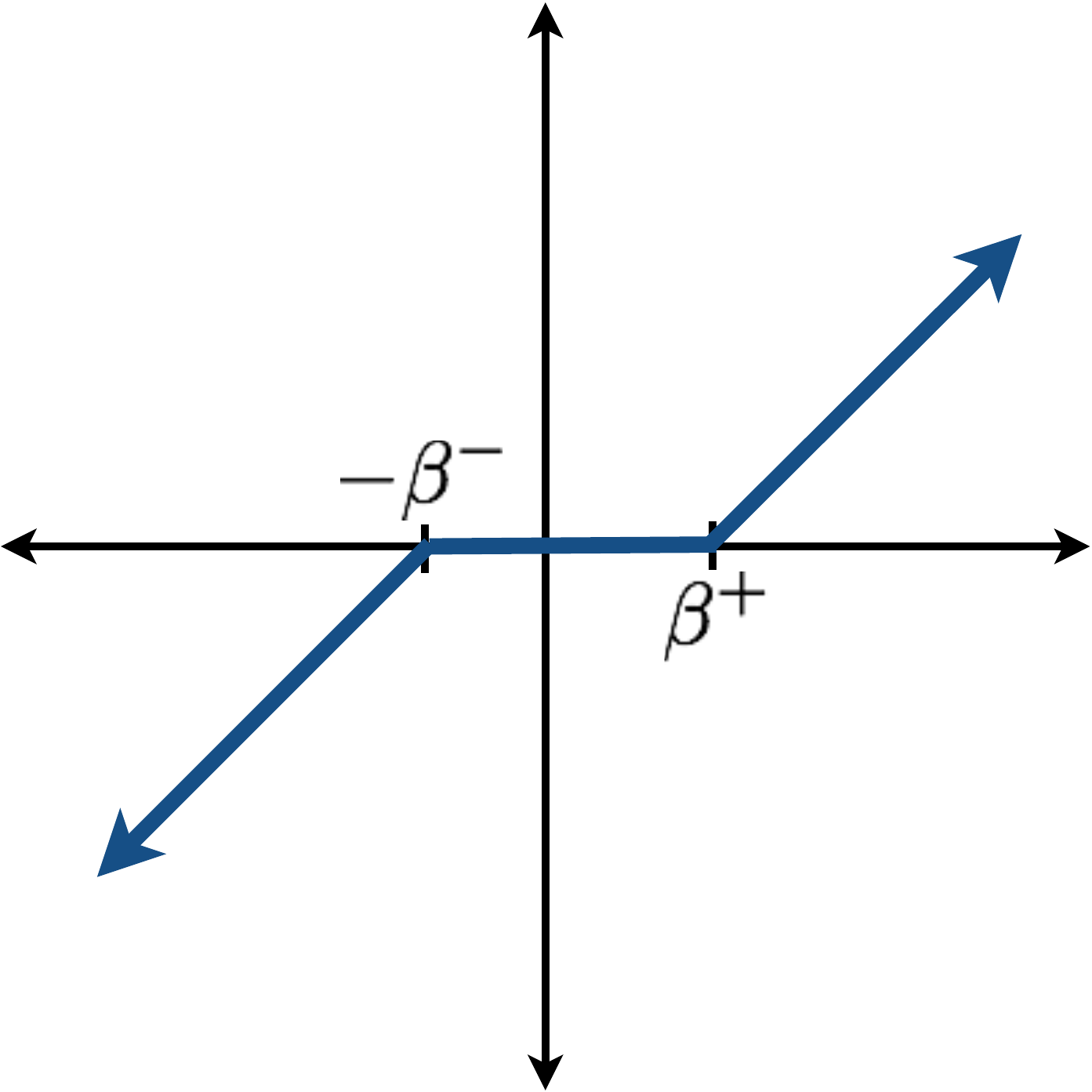}} &
  \subfloat[$0 \le -\beta^- \le \beta^+$]{\includegraphics[height=3cm]{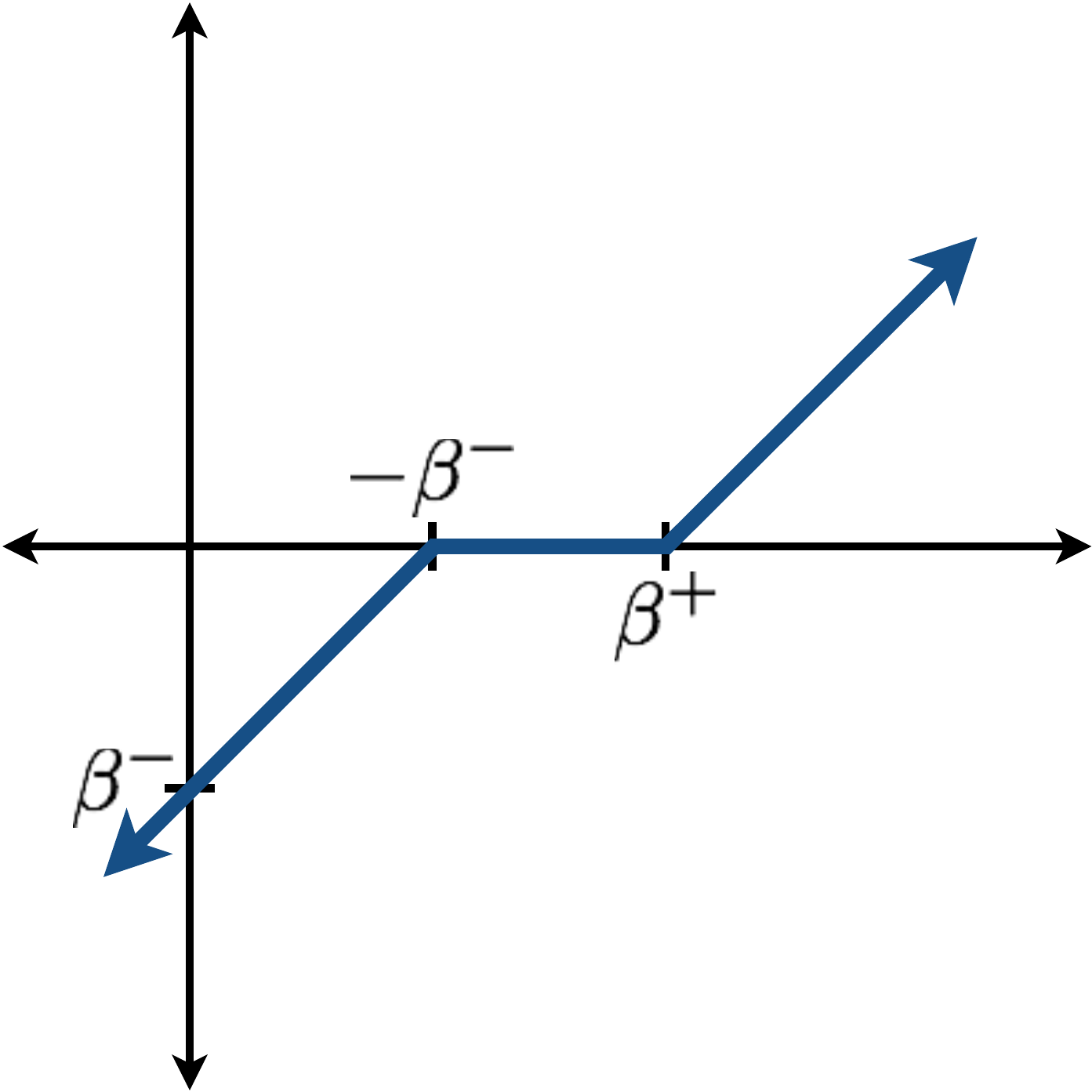}} &
  \subfloat[$-\beta^- \le \beta^+ \le 0$]{\includegraphics[height=3cm]{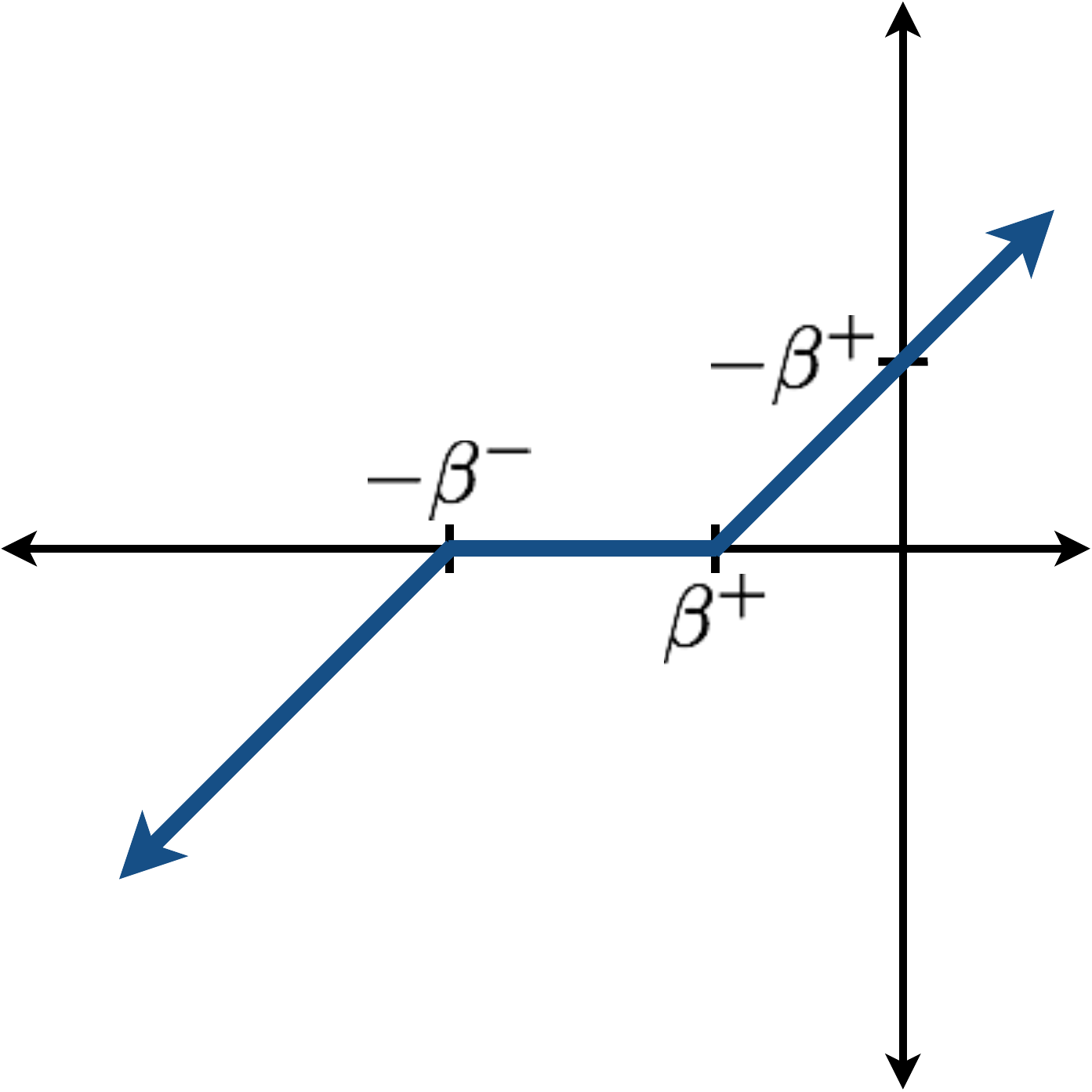}} &
  \subfloat[$\beta^- \le 0 \le -\beta^+$]{\includegraphics[height=3cm]{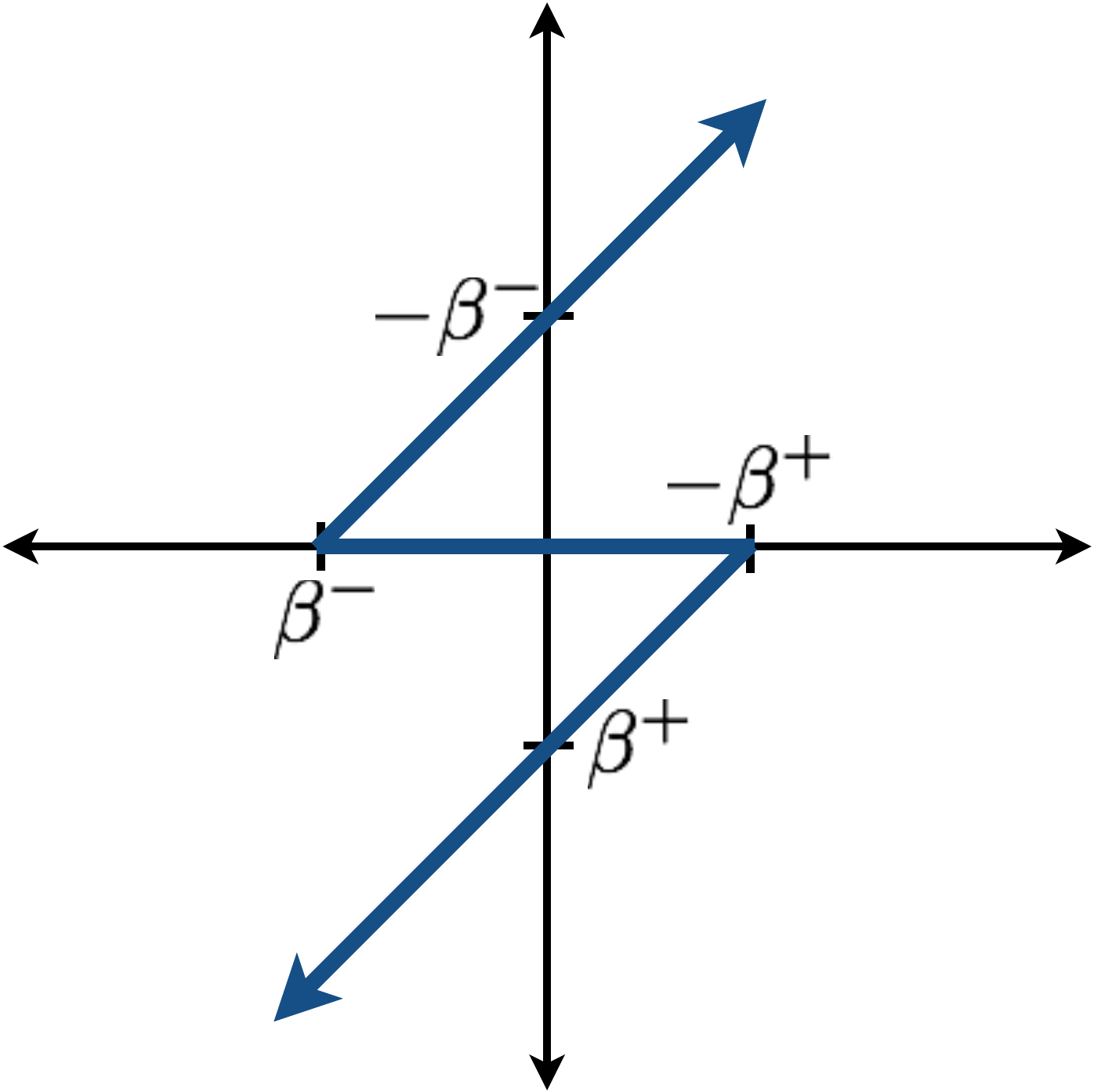}}
\end{tabular}
\caption{Comparing the behavior of asymmetric shrinkage for different settings
  of $\beta^+$ and $\beta^-$. (a)-(c) satisfy the condition that
  $-\beta^-\le\beta^+$ while (d) does not.}
\label{fig:shrink_scenarios}
\end{figure}

Fig.~\ref{fig:shrink_scenarios} shows a visualization of this function which
generalizes the standard shrinkage proximal operator by allowing for the
positive and negative thresholds.
In particular, it corresponds to the proximal operator for a version of the
$\ell_1$-norm that penalizes the positive and negative components with different
weights $|\vec{v}|_{asym} = \beta^+ \|\vec{v}^+\|_1 + \beta^- \|\vec{v}^-\|_1$.
The standard shrink operator corresponds to
$\operatorname{shrink}_{(\beta,-\beta)}(v)$ while the rectified linear unit
common in CNNs is given by a limiting case
$\operatorname{shrink}_{(0,-\infty)}(v)$.
We note that $-\beta^- \le \beta^+$ is required for
$\operatorname{shrink}_{(\beta^+,\beta^-)}$ to be a proper function (see
Fig.~\ref{fig:shrink_scenarios}).

\subsubsection{Feed-Forward Coding}
\label{sec:ff_coding}
We now describe how codes can be computed in a simple feed-forward pass.
Let
\begin{equation}
  {\symvec{\beta}}_{yk}^+=\beta-\vec{w}_{yk}^+, \qquad
  {\symvec{\beta}}_{yk}^-=\beta-\vec{w}_{yk}^-
  \label{eq:class_bias}
\end{equation}
be vectors of positive and negative biases whose entries are associated
with a spatial location in the feature map $k$ for class $y$.
The optimal code $\vec{z}$ can be computed in three sequential steps:
\begin{enumerate}
\item Cross-correlate the data with the filterbank $\vec{d}_k \star \vec{x}$
\item Apply an asymmetric version of the standard shrinkage operator
  \begin{equation}
    \tilde{\vec{z}}_k =
      \operatorname{shrink}_{({\symvec{\beta}}_{yk}^+,{\symvec{\beta}}_{yk}^-)}
        (\vec{d}_k \star \vec{x})
      \label{eq:ebm_direction}
  \end{equation}
where, with abuse of notation, we allow the shrinkage function
(Eq.~\ref{eq:asym_shrink}) to apply entries in the vectors of threshold
parameter pairs ${\symvec{\beta}}_{yk}^+,{\symvec{\beta}}_{yk}^-$ to the
corresponding elements of the argument.
\item Project onto the feasible set of unit length codes
\begin{equation}
  \vec{z}^* =
    \frac{\tilde{\vec{z}}}
         {\|\tilde{\vec{z}}\|_2}.
    \label{eq:ebm_projection}
\end{equation}
\end{enumerate}

\subsubsection{Relationship to CNNs:}
We note that this formulation of coding has a close connection to single layer
convolutional neural network (CNN).
A typical CNN layer consists of convolution with a filterbank followed by a
non-linear activation such as a rectified linear unit (ReLU).
ReLUs can be viewed as another way of inducing sparsity, but rather than coring
the values around zero like the shrink function, ReLU truncates negative values.
On the other hand, the asymmetric shrink function can be viewed as the sum of
two ReLUs applied to appropriately biased inputs:
\begin{equation*}
  \operatorname{shrink}_{(\beta^+,\beta^-)}(x) =
  \operatorname{ReLU}(x-\beta^+)-\operatorname{ReLU}(-(x+\beta^-)),
\end{equation*}
SSC coding can thus be seen as a CNN in which the ReLU activation has been
replaced with shrinkage followed by a global normalization.

\section{Learning}
We formulate supervised learning using the softmax log-loss that maximizes the
energy for the true class label $y_i$ while minimizing energy of incorrect labels $\bar{y}$.
\begin{equation}
\begin{split}
  \underset{\vec{d},\vec{w}^+,\vec{w}^-,\beta\ge0}{\arg\min}\, &
    \frac{\alpha}{2} (\|\vec{w}^+\|_2^2 + \|\vec{w}^-\|_2^2 + \|\vec{d}\|^2_2) \\
    &\!\!\!+\frac{1}{N}\sum_{i=1}^N
    [ -\max_{\|\vec{z}\|_2\le1} E(\vec{x}_i,y_i,\vec{z})
      +\log \sum_{\bar{y}\in\mathcal{Y}}
        \max_{\|\bar{\vec{z}}\|_2\le1} e^{E(\vec{x}_i,\bar{y},\bar{\vec{z}})}
    ] \\
    \text{s.t. }\,
    %& (\beta-\transpose{\mat{P}}\vec{w}_{yk}^+) + (\beta-\transpose{\mat{P}}\vec{w}_{yk}^-) \leq 0 \quad \forall y,k
    & 
    -(\beta-\vec{w}_{yk}^-) \leq (\beta-\vec{w}_{yk}^+) \quad \forall y,k
\end{split},
\label{eq:train_obj}
\end{equation}
where $\alpha$ is the hyperparameter regularizing $\vec{w}_y^+$, $\vec{w}_y^-$,
and $\vec{d}$.
We constrain the relationship between $\beta$ and the entries of $\vec{w}_y^+$
and $\vec{w}_y^-$ in order for the asymmetric shrinkage to be a proper function
(see Sec.~\ref{sec:asym} and Appendix B for details).

In classical sparse coding, it is typical to constrain the $\ell_2$-norm of
each dictionary filter to unit length.  Our spherical coding objective behaves
similarly.  For any optimal code $\vec{z}^*$, there is a $1$-dimensional
subspace of parameters for which $\vec{z}^*$ is optimal given by scaling
$\vec{d}$ inversely to $\vec{w}$, $\beta$.  For simplicity of the implementation,
we opt to regularize $\vec{d}$ to assure a unique solution.  However, as
\citet{tygert2015convolutional} point out, it may be advantageous from the
perspective of optimization to explicitly constrain the norm of the filterbank.

Note that unlike classical sparse coding, where $\beta$ is a hyperparameter
that is usually set using cross-validation, we treat it as a parameter of the
model that is learned to maximize performance.

\subsection{Optimization}

%Our formulation (Eq.~\ref{eq:train_obj}) is bilinear in $\vec{d}$ and
%$\vec{z}$, and therefore not jointly convex.
%We take an alternating minimization approach to find its solution.
In order to solve Eq.~\ref{eq:train_obj}, we explicitly formulate our model as
a directed-acyclic-graph (DAG) neural network with shared weights, where the
forward-pass computes the sparse code vectors and the backward-pass updates the
parameter weights. We optimize the objective using stochastic gradient descent
(SGD).

As mentioned in Sec.~\ref{sec:coding} shrinkage function is asymmetric with
parameters $\symvec{\beta}_{yk}^+$ or $\symvec{\beta}_{yk}^-$ as defined in
Eq.~\ref{eq:class_bias}.
However, the inequality constraint on their relationship to keep the shrinkage
function a proper function is difficult to enforce when optimizing with SGD.
Instead, we introduce a central offset parameter and reduce the ordering constraint
to pair of positivity constraints.
Let
\begin{equation}
  \hat{\vec{w}}_{yk}^+=\symvec{\beta}_{yk}^+ -b_k \qquad
  \hat{\vec{w}}_{yk}^-=\symvec{\beta}_{yk}^- +b_k
  \label{eq:class_bias2}
\end{equation}
be the modified linear ``classifiers'' relative to the central offset $b_k$.
It is straightforward to see that if $\symvec{\beta}_{yk}^+$ and
$\symvec{\beta}_{yk}^-$ that satisfy the constrain in Eq.~\ref{eq:train_obj},
then adding the same value to both sides of the inequality will not change that.
However, taking $b_k$ to be a midpoint between them, then both
$\symvec{\beta}_{yk}^+ -b_k$ and $\symvec{\beta}_{yk}^- +b_k$ will be strictly
non-negative.  %---which is why we can simplify the constraint.

Using this variable substitution, we rewrite the energy function
(Eq.~\ref{eq:energy_fn}) as
\begin{equation}
  E^\prime(\vec{x},y,\vec{z}) =
    \transpose{\vec{x}}\big( \sum_{k=1}^K \vec{d}_k*\vec{z}_k \big)
    +\sum_{k=1}^K b_k\transpose{\vec{1}}\vec{z}_k
    -\sum_{k=1}^K \hat{\vec{w}}_{yk}^{+\intercal}\vec{z}_k^+
    +\sum_{k=1}^K \hat{\vec{w}}_{yk}^{-\intercal}\vec{z}_k^-.
\label{eq:energy_fn2}
\end{equation}
where ${\vec b}$ is constant offset for each code channel.
The modified linear ``classification'' terms now takes on a dual role of inducing
sparsity and measuring the compatibility between $\vec{z}$ and $y$.

This yields a modified learning objective that can easily be solved with existing
implementations for learning convolutional neural networks:
\begin{equation}
\begin{split}
  \underset{\vec{d},\hat{\vec{w}}^+,\hat{\vec{w}}^-,\vec{b}}{\arg\min}\, &
    \frac{\alpha}{2} (\|\hat{\vec{w}}^+\|_2^2 + \|\hat{\vec{w}}^-\|_2^2 + \|\vec{d}\|^2_2) \\
    &\!\!\!+\frac{1}{N}\sum_{i=1}^N
    [ -\max_{\|\vec{z}\|_2\le1} E^\prime(\vec{x}_i,y_i,\vec{z})
      +\log \sum_{\bar{y}\in\mathcal{Y}}
        \max_{\|\bar{\vec{z}}\|_2\le1} e^{E^\prime(\vec{x}_i,\bar{y},\bar{\vec{z}})}
    ] \\
    \text{s.t. }\, & \hat{\vec{w}}_{yk}^+, \hat{\vec{w}}_{yk}^- \succeq 0 \quad
    \forall y,k
\end{split},
\label{eq:train_obj2}
\end{equation}
where $\hat{\vec{w}}^+$ and $\hat{\vec{w}}^-$ are the new sparsity inducing
classifiers, and $\vec{b}$ are the arbitrary origin points.
In particular, adding the $K$ origin points allows us to enforce the constraint
by simply projecting $\hat{\vec{w}}^+$ and $\hat{\vec{w}}^-$ onto the positive
orthant during SGD.

\subsubsection{Stacking Blocks}
\label{sec:stack_coding}
We also examine stacking multiple blocks of our energy function in order to
build a hierarchical representation.  As mentioned in Sec.~\ref{sec:ff_coding},
the optimal codes can be computed in a simple feed-forward pass---this applies
to shallow versions of our model.  When stacking multiple blocks of our
energy-based model, solving for the optimal codes cannot be done in a
feed-forward pass since the codes for different blocks are coupled (bilinearly)
in the joint objective.  Instead, we can proceed in an iterative manner,
performing block-coordinate descent by repeatedly passing up and down the
hierarchy updating the codes.  In this section we investigate the trade-off
between the number of passes used to find the optimal codes for the stacked
model and classification performance.

For this purpose, we train multiple instances of a 2-block version of our
energy-based model that differ in the number of iterations used when solving
for the codes.  For recurrent networks such as this, inference is commonly
implemented by ``unrolling'' the network, where the parts of the network
structure are repeated with parameters shared across these repeated parts to
mimic an iterative algorithm that stops at a fixed number of iterations rather
than at some convergence criteria.

\begin{figure}[h]
\centering
\begin{tabular}{cc}
  \subfloat[Train Objective]{\includegraphics[width=2.6in]{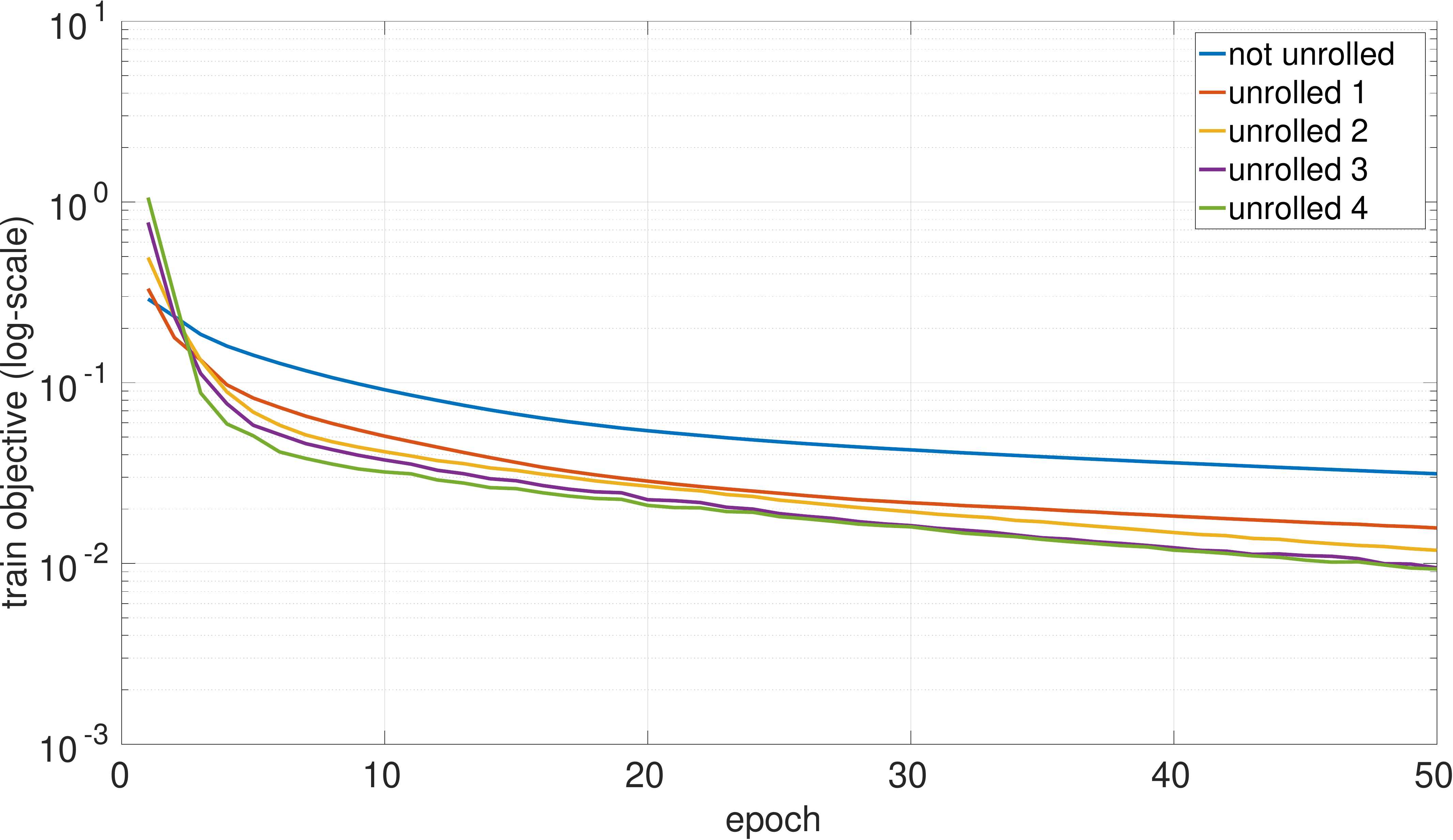}} &
  \subfloat[Test Error]{\includegraphics[width=2.6in]{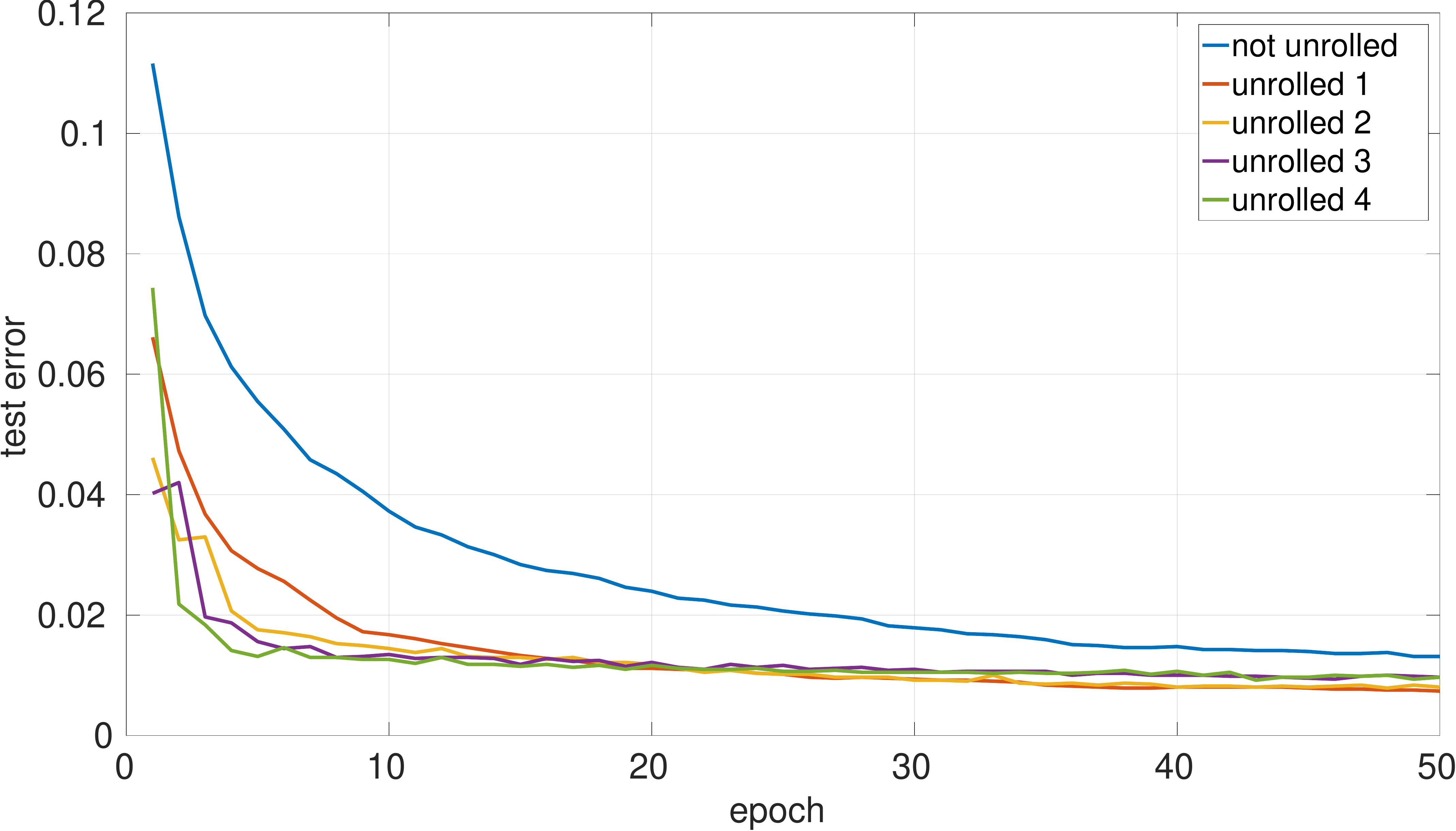}}
\end{tabular}
\caption{Comparing the effects of unrolling a 2-block version of our
energy-based model. (Best viewed in color.)}
\label{fig:unrolled}
\end{figure}

In Fig.~\ref{fig:unrolled}, we compare the performance between models that were
unrolled zero to four times.  We see that there is a difference in performance
based on how many sweeps of the variables are made.  In terms of the training
objective, more unrolling produces models that have lower objective values with
convergence after only a few passes.  In terms of the testing error rate, however, we
see that full code inference is not necessarily better, as unrolling once or
twice has lower error rates than unrolling three or four times.  The biggest
difference was between not unrolling and unrolling once, where both the
training objective and testing error rate go down.  The testing error rate decreases
from 0.0131 to 0.0074.  While there is a clear benefit in terms of performance
for unrolling at least once, there is also a trade-off between performance and
computational resource, especially for deeper models.

\section{Experiments}
We evaluate the benefits of combining top-down and bottom-up information to
produce class-specific features on the CIFAR-10~\citep{krizhevsky2009learning}
dataset using a deep version of our EB-SSC.
All experiments were performed using the
{MatConvNet}~\citep{vedaldi15matconvnet} framework with the ADAM
optimizer~\citep{kingma2014adam}.
The data was preprocessed and augmented following the procedure in
\citet{goodfellow2013maxout}.
Specifically, the data was made zero mean and whitened, augmented with
horizontal flips (with a 0.5 probability) and random cropping.  No weight decay
was used, but we used a dropout rate of $0.3$ before every convolution layer
except for the first. For these experiments we consider a single forward pass
(no unrolling).

\begin{table}
\centering
%\begin{minipage}{0.495\textwidth}
\begin{tabular}{c|c|c}
  \hline
  \multicolumn{3}{c}{Base Network} \\\hline
  block & kernel, stride, padding & activation \\\hline
  conv1 & $3\times3\times3\times96, 1, 1$ & ReLU/CReLU \\\hline
  conv2 & $3\times3\times96/192\times96, 1, 1$ & ReLU/CReLU \\\hline
  pool1 & $3\times3, 2, 1$ & max \\\hline
  conv3 & $3\times3\times96/192\times192, 1, 1$ & ReLU/CReLU \\\hline
  conv4 & $3\times3\times192/384\times192, 1, 1$ & ReLU/CReLU \\\hline
  conv5 & $3\times3\times192/384\times192, 1, 1$ & ReLU/CReLU \\\hline
  pool2 & $3\times3, 2, 1$ & max \\\hline
  conv6 & $3\times3\times192/384\times192, 1, 1$ & ReLU/CReLU \\\hline
  conv7 & $1\times1\times192/384\times192, 1, 1$ & ReLU/CReLU \\\hline\hline
\end{tabular}
%\end{minipage}
%
%\begin{minipage}{0.495\textwidth}
%\begin{tabular}{c|c|c}
%  \hline
%  & \multicolumn{2}{c}{Linear Classifier} \\\hline
%  block & kernel, stride, padding & activation \\\hline
%  conv8 & $1\times1\times384\times10, 1, 0$ & ReLU \\\hline
%  pool3 & $10\times10, 1, 0$ & avg \\\hline\hline
%\end{tabular}
%
%\begin{tabular}{c|c|c|c}
%  \hline
%  & \multicolumn{3}{c}{Energy-Based Classifier} \\\hline
%  block & kernel, stride, padding & activation & normalization \\\hline
%  classbias & $1\times1, 1, 0$ & CReLU & SphNorm \\\hline
%\end{tabular}
%\end{minipage}
\caption{Underlying block architecture common across all models we evaluated.
SSC networks add an extra normalization layer after the non-linearity.
And EB-SSC networks insert class-specific bias layers between the convolution
layer and the non-linearity. Concatenated ReLU (CReLU) splits positive and 
negative activations into two separate channels rather than discarding the 
negative component as in the standard ReLU.}
\label{tab:model_def}
\end{table}

\subsection{Classification}
We compare our proposed EB-SSC model to that of
\citet{springenberg2015striving}, which uses rectified linear units (ReLU) as
its non-linearity.  This model can be viewed as a basic feed-forward version of
our proposed model which we take as a baseline.  We also consider variants of
the baseline model that utilize a subset of architectural features of our
proposed model (\eg concatenated rectified linear units (CReLU) and spherical
normalization (SN)) to understand how subtle design changes of the network
architecture affects performance.

We describe the model architecture in terms of the feature extractor and classifier.
Table~\ref{tab:model_def} shows the overall network architecture of feature
extractors, which consist of seven convolution blocks and two pooling layers.
%We considered max-pooling and average-pooling for feature extraction and indicate
%which operation in the subscript of the feature extractor (\eg SSC$_{max}$
%is used to denote average-pooling with spherical sparse coding features).
We test two possible classifiers: a simple linear classifier (LC) and our
energy-based classifier (EBC), and use softmax-loss for all models.
For linear classifiers, a numerical subscript indicates which of the seven conv
blocks of the feature extractor is used for classification (\eg LC$_7$ indicates
the activations out of the last conv block is fed into the linear classifier).
For energy-based classifiers, a numerical subscript indicates which conv blocks
of the feature extractor are replace with a energy-based classifier (\eg
EBC$_{6-7}$ indicates the activations out of conv5 is fed into the energy-based
classifier and the energy-based classifier has a similar architecture to the
conv blocks it replaces).  The notation differ because for energy-based
classifiers, the optimal activations are a function of the hypothesized class
label, whereas for linear classifiers, they are not.

\begin{table}[h]
\centering
\begin{tabular}{l|c|c|c}
  \hline
  Model & Train Err. (\%) & Test Err. (\%) & \# params \\\hline
  ReLU+LC$_7$ & 1.20 & 11.40 & 1.3M \\\hline
%  CReLU$_{avg}$+LC$_7$ & 1.16 & 11.01 & 2.6M \\\hline
  CReLU+LC$_7$ & 2.09 & 10.17 & 2.6M \\\hline
%  CReLU$_{max}$+EBC$_7$ & 0.63 & 10.49 & 2.9M \\\hline
  CReLU(SN)+LC$_7$ & 0.99 & 9.74 & 2.6M \\\hline
%  SSC$_{avg}$+LC$_7$ & 1.52 & 11.84 & 2.6M \\\hline
  SSC+LC$_7$ & 0.99 & 9.77 & 2.6M \\\hline
%  SSC$_{avg}$+EBC$_7$ & 3.91 & 13.64 & 2.9M \\\hline
%  SSC$_{avg}$+EBC$_{1-7}$ & 3.24 & 15.62 & 10.1M \\\hline
%  SSC+LC$_7$ (unrolled) & 3.27 & 13.41 & 2.6M \\\hline
  SSC+EBC$_{6-7}$ & 0.21 & 9.23 & 3.2M \\\hline
\end{tabular}
\caption{Comparison of the baseline ReLU+LC$_7$ model, its derivative
  models, and our proposed model on CIFAR-10.}
\label{tab:cifar10_perf}
\end{table}

The results shown in Table~\ref{tab:cifar10_perf} compare our proposed model to
the baselines ReLU+LC$_7$~\citep{springenberg2015striving} and
CReLU+LC$_7$~\citep{shang2016understanding}, and to
intermediate variants.  The baseline models all perform very similarly with 
some small reductions in error rates over the baseline CReLU+LC$_7$.
However, CReLU+LC$_7$ reduces the error rate over ReLU+LC$_7$ by more than one
percent (from 11.40\% to 10.17\%), which confirms the claims by
\citet{shang2016understanding} and demonstrates the benefits of splitting
positive and negative activations.
Likewise, we see further decrease in the error rate (to 9.74\%) from using
spherical normalization.
Though normalizing the activations doesn't add any capacity to the model, this
improved performance is likely because scale-invariant activations makes
training easier.
On the other hand, further sparsifying the activations yielded no benefit.
We tested values $\beta=\{0.001, 0.01\}$ and found $0.001$ to perform better.
Replacing the linear classifier with our energy-based classifier further
decreases the error rate by another half percent (to 9.23\%).
%Our full energy based model performs substantially worse compared to the
%baseline model, (15.62\% versus 10.68\%).  There are two factors that may be
%contributing to this worse performance.  The first is that we did no
%optimization of learning hyperparameters (learning rates and dropout rates).
%Evaluation was done using the same parameters across the board and were obtained
%from \citet{shang2016understanding}.  Given that the full $EBC_{1-7}$ model has
%substantially higher capacity (10.1M parameters versus 2.6M parameters), we
%expect that the model tested here is under-fit (as suggested by the training
%error).  The second factor is that, as can be seen in the table,
%average-pooling generally performs worse than max pooling in other models.  We
%expect the same to be true here.  As of the submission deadline the max-pooled
%variants were still training.

\subsection{Decoding Class-Specific Codes}
A unique aspect of our model is that it is generative in the sense that each
layer is explicitly trying to encode the activation pattern in the prior layer.
Similar to the work on deconvolutional networks built on least-squares sparse
coding~\citep{zeiler_deconvolutional_2010}, we can synthesize input images 
from activations in our spherical coding network by performing repeated
deconvolutions (transposed convolutions) back through the network.  Since our
model is energy based, we can further examine how the top-down information of a
hypothesized class effects the intermediate activations.

\begin{figure}[h]
\centering
\includegraphics[width=2.7in]{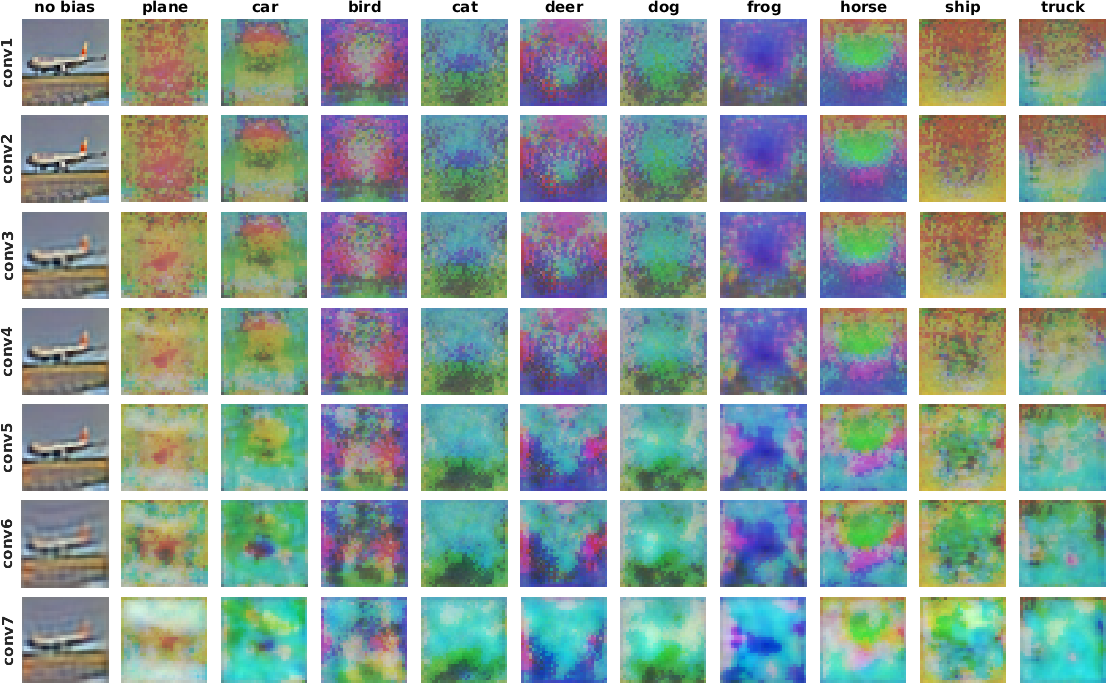}
\caption{The reconstruction of an airplane image from different levels of the
network (rows) across different hypothesized class labels (columns).
The first column is pure reconstruction, \ie unbiased by a hypothesized class
label, the remaining columns show reconstructions of the learned class bias
at each layer for one of ten possible CIFAR-10 class labels.
(Best viewed in color.)}
%\includegraphics[width=2.7in]{mnist_blocks_vs_class}
%\caption{The reconstruction of a digit eight image from different levels of the
%network (rows) across different hypothesized class labels (columns).
%The first column is pure reconstruction, \ie unbiased by a hypothesized class
%label, the remaining columns are one of ten possible MNIST class labels.}
\label{fig:decode}
\end{figure}

The first column in Fig.~\ref{fig:decode} visualizes reconstructions of a 
given input image based on activations from different layers of the model by
convolution transpose. In this case we put in zeros for class biases (\ie no top-down) and
are able to recover high fidelity reconstructions of the input.  In the 
remaining columns, we use the same deconvolution pass to construct input space
representations of the learned classifier biases.  At low levels of the feature
hierarchy, these biases are spatially smooth since the receptive fields are
small and there is little spatial invariance capture in the activations.
At higher levels these class-conditional bias fields become more tightly
localized.

\begin{figure}[h]
\centering
\subfloat[conv2]{%
  \begin{minipage}[c]{0.5\textwidth}
    \centering
    \includegraphics[width=2.7in]{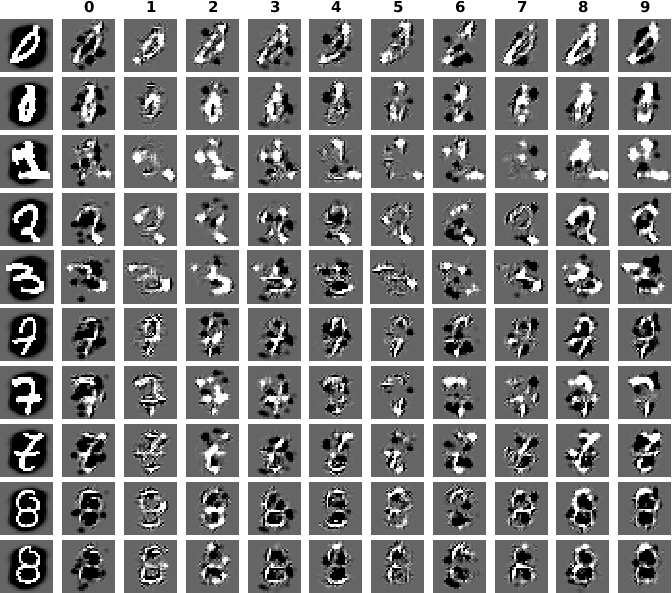}
  \end{minipage}}
\subfloat[conv5]{%
  \begin{minipage}[c]{0.5\textwidth}
    \centering
    \includegraphics[width=2.7in]{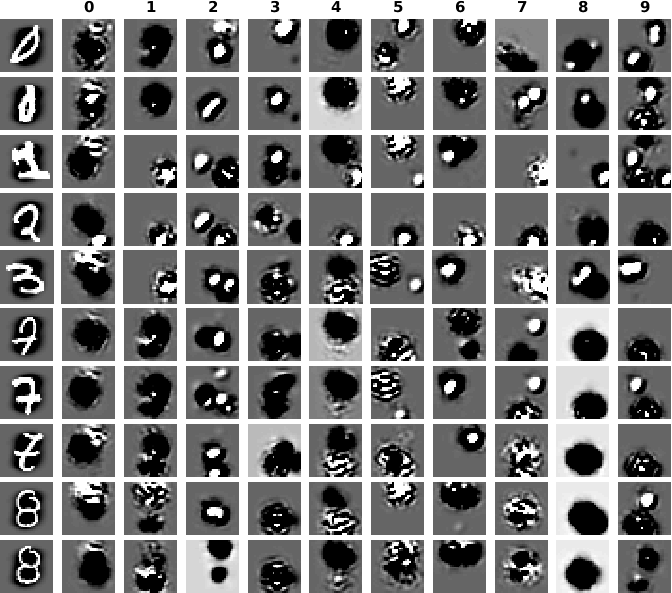}
  \end{minipage}}
\caption{Visualizing the reconstruction of different input images (rows)
for each of 10 different class hypotheses (cols) from the 2nd and 5th block
activations for a model trained on MNIST digit classification.}
\label{fig:decode_residual}
\end{figure}

Finally, in Fig.~\ref{fig:decode_residual} we shows decodings from the conv2
and conv5 layer of the EB-SSC model for a given input under different class
hypotheses. Here we subtract out the contribution of the top-down bias term in
order to isolate the effect of the class conditioning on the encoding of input
features. As visible in the figure, the modulation of the activations focused
around particular regions of the image and the differences across class 
hypotheses becomes more pronounced at higher layers of the network.

\section{Conclusion}
We presented an energy-based sparse coding method that efficiently combines
cosine similarity, convolutional sparse coding, and linear classification.
Our model shows a clear mathematical connection between the activation functions
used in CNNs to introduce sparsity and our cosine similarity convolutional
sparse coding formulation.
Our proposed model outperforms the baseline model and we show which attributes
of our model contributes most to the increase in performance.
We also demonstrate that our proposed model provides an interesting framework to
probe the effects of class-specific coding.

%\clearpage
\bibliography{refs}
\bibliographystyle{iclr2017_conference}
\clearpage

%-------------------------------------------------------------------------
\section*{Appendix A}
\label{app:deriv_ipcsc}
Here we show that spherical sparse coding (SSC) with a norm constraint on the
reconstruction is equivalent to standard convolutional sparse coding (CSC).
Expanding the least squares reconstruction error and dropping the constant
term $\|x\|_2^2$ gives the CSC problem:
\begin{equation*}
  \max_\vec{z} \,\, 2\transpose{\vec{x}}\big(
  \sum_{k=1}^K\vec{d}_k*\vec{z}_k \big)
  -\|\sum_{k=1}^K\vec{d}_k*\vec{z}_k\|_2^2 -\beta\sum_{k=1}^K\|\vec{z}_k\|_1.
\end{equation*}
\noindent Let $\epsilon = \|\sum_{k=1}^K\vec{d}_k*\vec{z}_k\|_2$ be the norm of
the reconstruction for some code $\vec{z}$ and let $\vec{u}$ be the
reconstruction scaled $\epsilon$ to have unit norm so that:
\begin{equation*}
  \vec{u} = \frac{\sum_{k=1}^K\vec{d}_k*\vec{z}_k}{\|\sum_{k=1}^K\vec{d}_k*\vec{z}_k\|_2}
           = \sum_{k=1}^K\vec{d}_k*\bar{\vec{z}}_k
  \quad \text{with} \quad
  \bar{\vec{z}} = \frac{1}{\epsilon}\vec{z}
\end{equation*}
We rewrite the least-squares objective in terms of these new variables:
\begin{equation*}
\begin{split}
  \underset{\bar{\vec{z}},\epsilon>0}{\max}\,\,
  g(\bar{\vec{z}},\epsilon) &=\underset{\bar{\vec{z}},\epsilon>0}{\max}\,\, 2\transpose{\vec{x}}\big( \epsilon\vec{u} \big)
  -\|\epsilon\vec{u}\|_2^2 -\beta \|\epsilon \bar{\vec{z}}\|_1 \\
  &= \underset{\bar{\vec{z}},\epsilon>0}{\max}\,\,2\epsilon\big( \transpose{\vec{x}}\vec{u}
  -\frac{\beta}{2} \|\bar{\vec{z}}\|_1 \big) -\epsilon^2
\end{split}
\end{equation*}
Taking the derivative of $g$ \wrt $\epsilon$ yields the optimal scaling
$\epsilon^*$ as a function of $\bar{\vec{z}}$:
\begin{equation*}
  \epsilon(\bar{\vec{z}})^* = \transpose{\vec{x}}\vec{u}
  -\frac{\beta}{2}\|\bar{\vec{z}}\|_1.
\end{equation*}
Plugging $\epsilon(\bar{\vec{z}})^*$ back into $g$ yields:
\begin{equation*}
  \underset{\bar{\vec{z}},\epsilon>0}{\max}\,\, g(\bar{\vec{z}},\epsilon) =
  \underset{\bar{\vec{z}},\|u\|_2=1}{\max}\,\, \big( \transpose{\vec{x}}\vec{u} -\frac{\beta}{2}\|\bar{\vec{z}}\|_1 \big)^2.
\end{equation*}
Discarding solutions with $\epsilon < 0$ can be achieved by simply
dropping the square which results in the final constrained problem:
\begin{equation*}
\begin{split}
  \underset{\bar{\vec{z}}}{\arg\max}\,\, &\transpose{\vec{x}}\big(
  \sum_{k=1}^K\vec{d}_k*\bar{\vec{z}}_k \big)
  -\frac{\beta}{2}\sum_{k=1}^K\|\bar{\vec{z}}_k\|_1 \\
  \text{s.t.}\quad &\|\sum_{k=1}^K\vec{d}_k*\bar{\vec{z}}_k\|_2 \le 1.
\end{split}
\end{equation*}

%-------------------------------------------------------------------------
\section*{Appendix B}
\label{app:coding_ipcsc}
We show in this section that coding in the EB-SSC model can be solved
efficiently by a combination of convolution, shrinkage and projection,
steps which can be implemented with standard libraries on a GPU.
For convenience, we first rewrite the objective in terms of cross-correlation
rather than convolution (\ie, $\transpose{\vec{x}}(\vec{d}_k*\vec{z}_k) =
\transpose{(\vec{d}_k\star\vec{x})}\vec{z}_k$).  For ease of understanding,
we first consider the coding problem when there is no classification term.
\begin{equation*}
  \vec{z}^* =
  \underset{\|\vec{z}\|_2^2\le1}{\arg\max}\,\,
  \transpose{\vec{v}}\vec{z} -\beta\|\vec{z}\|_1,
\end{equation*}
where $\vec{v} = \transpose{[\transpose{(\vec{d}_1\star\vec{x})}, \ldots,
\transpose{(\vec{d}_K\star\vec{x})}]}$.
Pulling the constraint into the objective, we get its Lagrangian function:
\begin{equation*}
  \L(\vec{z},\lambda) = \transpose{\vec{v}}\vec{z} -\beta\|\vec{z}\|_1
    +\lambda\big( 1-\|\vec{z}\|_2^2 \big).
\end{equation*}
From the partial subderivative of the Lagrangian \wrt $z_i$ we derive the
optimal solution as a function of $\lambda$;
and from that find the conditions in which the solutions hold, giving us:
\begin{equation}
  z_i(\lambda)^* = \frac{1}{2\lambda}\cdot\left\{
    \begin{array}{cl}
      v_i-\beta & \qquad v_i>\beta \\
      0 & \qquad \text{otherwise} \\
      v_i+\beta & \qquad v_i<\beta
    \end{array}
    \right..
    \label{eq:shrink_derv}
\end{equation}
This can also be compactly written as:
\begin{align}
  \vec{z}(\lambda)^* = \frac{1}{2\lambda}\tilde{\vec{z}}, \label{eq:z_lambda} \\
  \tilde{\vec{z}} = \vec{s}^2\odot\vec{v} - \beta\vec{s} \nonumber
\end{align}
where $\vec{s}=\operatorname{sign}(\vec{z}^*)\in\{-1,0,1\}^{|\vec{z}|}$ and
$\vec{s}^2=\vec{s}\odot\vec{s}\in\{0,1\}^{|\vec{z}|}$.
The sign vector of $\vec{z}^*$ can be determined without knowing $\lambda$,
as $\lambda$ is a Lagrangian multiplier for an inequality it must be
non-negative and therefore does not change the sign of the optimal solution.
Lastly, we define the squared $\ell_2$-norm of $\tilde{\vec{z}}$, a result that
will be used later:
\begin{align}
  \|\tilde{\vec{z}}\|_2^2 &= \transpose{\tilde{\vec{z}}}(\vec{s}^2\odot\vec{v})
    -\beta\transpose{\tilde{\vec{z}}}\vec{s} \nonumber\\
    &= \transpose{\tilde{\vec{z}}}\vec{v} -\beta\|\tilde{\vec{z}}\|_1.
    \label{eq:z_tilde_sq}
\end{align}

Substituting $\vec{z}(\lambda)^*$ back into the Lagrangian we get:
\begin{equation*}
  \L(\vec{z}(\lambda)^*,\lambda)
    = \frac{1}{2\lambda}\transpose{\vec{v}}\tilde{\vec{z}}
     -\frac{\beta}{2\lambda}\|\tilde{\vec{z}}\|_1
     +\lambda\big( 1-\frac{1}{4\lambda^2}\|\tilde{\vec{z}}\|_2^2 \big),
\end{equation*}
and the derivative \wrt $\lambda$ is:
\begin{equation*}
  \frac{\partial\L(\vec{z}(\lambda)^*}{\partial\lambda} =
  -\frac{1}{2\lambda^2}\transpose{\vec{v}}\tilde{\vec{z}}
  +\frac{\beta}{2\lambda^2}\|\tilde{\vec{z}}\|_1 +1
  +\frac{1}{4\lambda^2}\|\tilde{\vec{z}}\|_2^2.
\end{equation*}
Setting the derivative equal to zero and using the result
from Eq.~\ref{eq:z_tilde_sq}, we can find the optimal solution to $\lambda$:
\begin{align*}
  \lambda^2 &= \frac{1}{2}\transpose{\tilde{\vec{z}}}\vec{v}
    -\frac{\beta}{2}\|\tilde{\vec{z}}\|_1
    -\frac{1}{4}\|\tilde{\vec{z}}\|_2^2
    = \frac{1}{2}\|\tilde{\vec{z}}\|_2^2 -\frac{1}{4}\|\tilde{\vec{z}}\|_2^2 \\
  \implies \lambda^* &= \frac{1}{2}\|\tilde{\vec{z}}\|_2.
\end{align*}
Finally, plugging $\lambda^*$ into Eq.~\ref{eq:z_lambda} we find the optimal
solution
\begin{equation}
  \vec{z}^* = \frac{\tilde{\vec{z}}}{\|\tilde{\vec{z}}\|_2}.
  \label{eq:z_opt}
\end{equation}

\end{document}